\pdfoutput=1

\documentclass[11pt]{article}

\usepackage[final]{acl}

\usepackage{times}
\usepackage{latexsym}
\usepackage[T1]{fontenc}
\usepackage[utf8]{inputenc}
\usepackage[nopatch=footnote]{microtype}
\usepackage{inconsolata}
\usepackage{graphicx}
\usepackage{booktabs}
\usepackage{setspace}
\usepackage{enumitem}
\usepackage{xcolor}
\usepackage{soul}
\usepackage{subfig} %
\usepackage{subcaption}

\usepackage{hyperref}
\usepackage{url}
\usepackage{multirow}
\usepackage[normalem]{ulem}
\usepackage{listings}   %

\usepackage{fdsymbol}   %
\usepackage{xspace}

\usepackage{tcolorbox}
\usepackage{pifont}%

\newtcolorbox{shadedlisting}{
    colback=gray!10,
    boxrule=0pt,
    left=15pt,
    right=15pt,
    top=10pt,
    bottom=10pt
}

\title{\tableIcon\textsc{arxivDIGESTables}: Synthesizing Scientific Literature into \\ Tables using Language Models}

\author{Benjamin Newman$^{\spadesuit}$\thanks{Equal contributions.} \quad Yoonjoo Lee$^{\heartsuit*}$ \\ 
\textbf{Aakanksha Naik}$^\diamondsuit$ \quad \textbf{Pao Siangliulue}$^\diamondsuit$ \quad \textbf{Raymond Fok}$^\spadesuit$ \\ 
\textbf{Juho Kim}$^\heartsuit$ \quad \textbf{Daniel S. Weld}$^{\spadesuit\diamondsuit}$ \quad \textbf{Joseph Chee Chang}$^\diamondsuit$ \quad \textbf{Kyle Lo}$^\diamondsuit$
\vspace{0.5em}
\\
$^\spadesuit$University of Washington \quad
$^\heartsuit$KAIST \quad
$^\diamondsuit$Allen Institute for AI
\vspace{0.5em}
\\
\texttt{blnewman@cs.washington.edu, yoonjoo.lee@kaist.ac.kr}
}

\newcommand{\xmark}{\ding{55}}%

\newcommand{\ourdata}{{\tableIcon\textsc{arxivDIGESTables}{}}}
\newcommand{\ourmetric}{{\textsc{DecontextEval}{}}}
\newcommand{\numtotaltables}{{{2,228}{}}}
\newcommand{\numtotalpapers}{{7,542}{}}

\newcommand{\hlc}[2][yellow]{{%
    \colorlet{foo}{#1}%
    \sethlcolor{foo}\hl{#2}}%
}
\definecolor{lightblue}{RGB}{220, 240, 255}
\definecolor{lightgreen}{RGB}{200, 255, 200}

\newcommand{\tableIcon}{\raisebox{-2pt}{\includegraphics[height=1em]{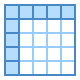}}}

\newcommand{\huggingface}{\raisebox{-1.5pt}{\includegraphics[height=1.05em]{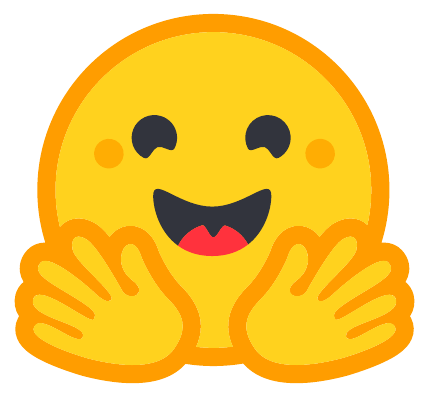}}\xspace}
\newcommand{\github}{\raisebox{-1.5pt}{\includegraphics[height=1.05em]{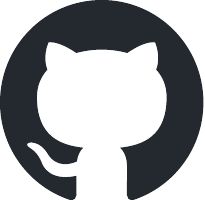}}\xspace}

\begin{document}
\maketitle

\begin{abstract}
When conducting literature reviews, scientists often create \textit{literature review tables}---tables whose rows are publications and whose columns constitute a \textit{schema}, a set of aspects used to compare and contrast the papers.
Can we automatically generate these tables using language models (LMs)? 
In this work, we introduce a framework that leverages LMs to perform this task by decomposing it into separate \emph{schema} and \emph{value} generation steps. 
To enable experimentation, we address two main challenges:
First, we overcome a lack of high-quality datasets to benchmark table generation by curating and releasing \ourdata{}, a new dataset of 2,228 literature review tables extracted from ArXiv papers that synthesize a total of 7,542 research papers.
Second, to support scalable evaluation of model generations against human-authored reference tables, we develop \ourmetric{}, an automatic evaluation method that aligns elements of tables with the same underlying aspects despite differing surface forms.
Given these tools, we evaluate LMs' abilities to reconstruct reference tables, finding this task benefits from additional context to ground the generation (e.g. table captions, in-text references).
Finally, through a human evaluation study we find that even when LMs fail to fully reconstruct a reference table, their generated novel aspects can still be useful.

\renewcommand{\arraystretch}{1.2}
\begin{tabular}{rl}
 \huggingface 
 & \href{https://huggingface.co/datasets/blnewman/arxivDIGESTables}{\path{blnewman/arxivDIGESTables}}\\
 \github & \href{https://github.com/bnewm0609/arxivDIGESTables}{\path{bnewm0609/arxivDIGESTables}} \\
\end{tabular}

\end{abstract}

\section{Introduction}
\label{sec:intro}

\begin{figure}[t]
    \centering
    \includegraphics[width=1.0\linewidth,trim={0cm 0cm 0cm 0cm},clip]{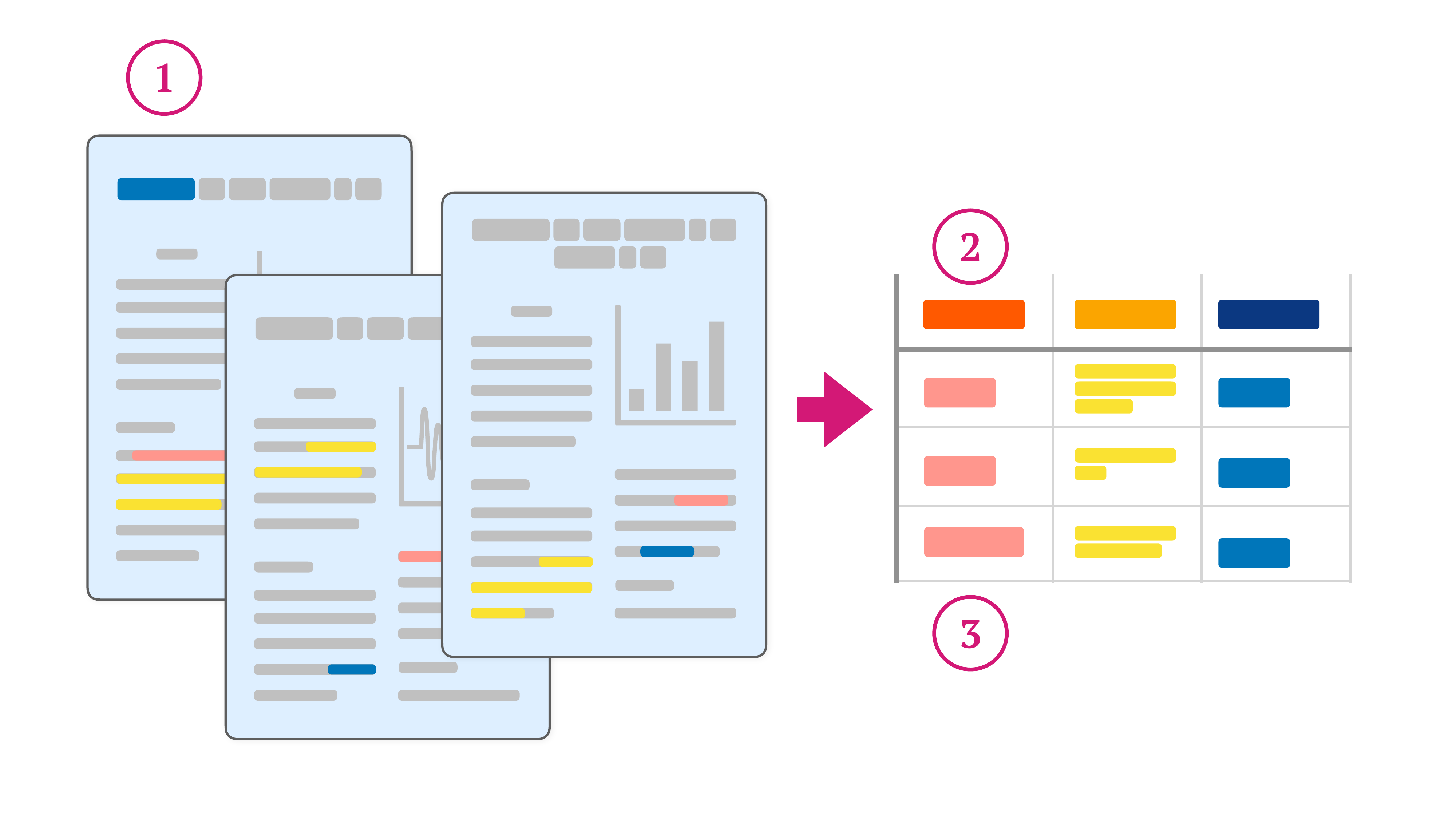}
    \caption{Schematic of our literature review table generation task: (1) synthesize multiple input papers into a table with both (2) a schema (columns) and (3) values. Each row corresponds to an input paper.}
    \label{fig:task-format}
\end{figure}

\begin{figure*}[t]
    \centering
    \includegraphics[width=1.0\textwidth,trim={3cm 0cm 3cm 0cm},clip]{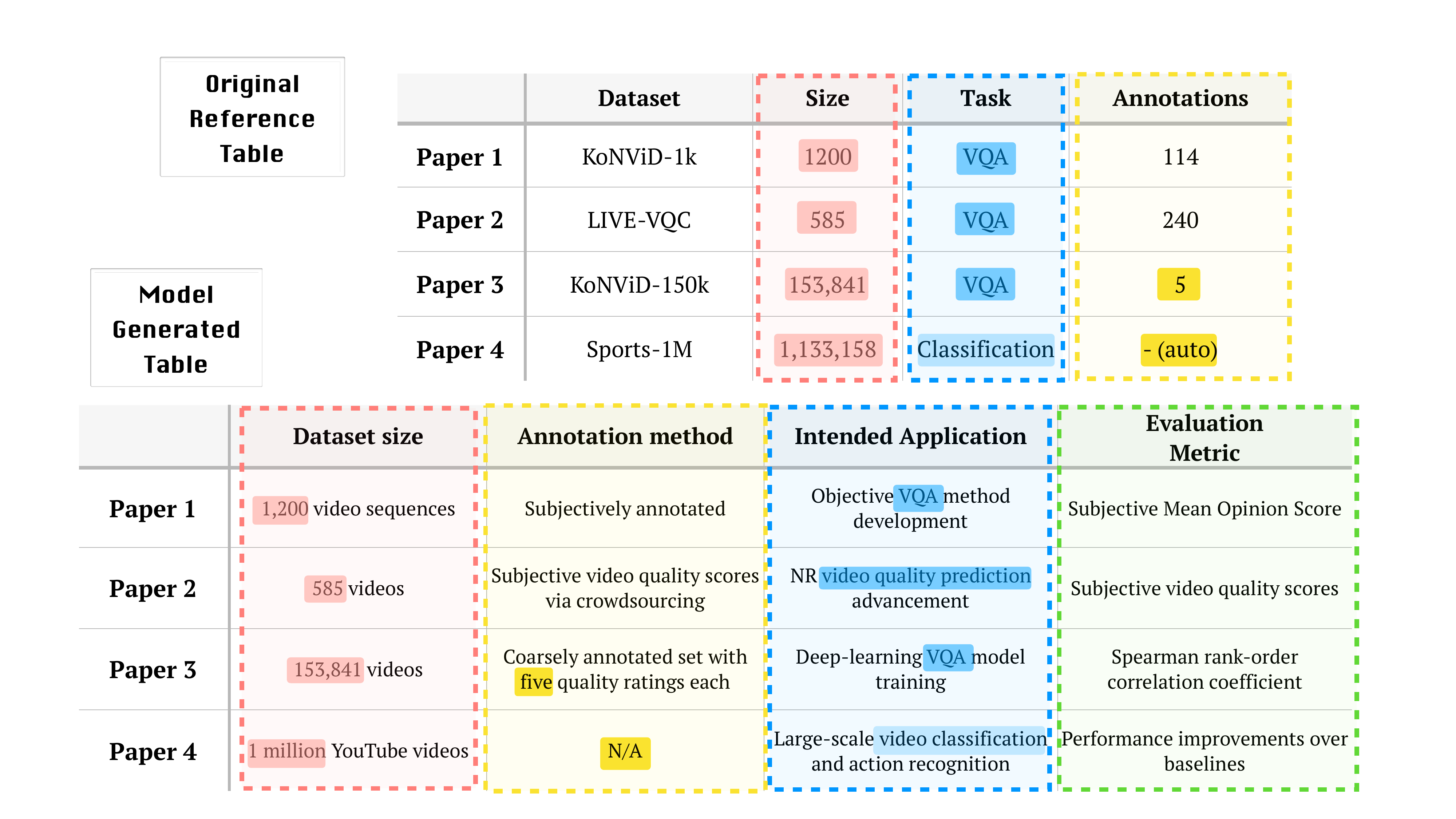}%
    \caption{Side-by-side comparison of a reference literature review table from an ArXiv paper \citep{LiuAda2023} and a model-generated table given the same input papers. 
    The generated table has reconstructed two gold aspects:
    the \hlc[pink!60]{pink} and \hlc[lightblue]{blue} aspects are the same, despite surface form differences (e.g., ``Task'' vs ``Intended Application'').
    The generated table has also proposed two novel aspects that are still relevant and useful, like ``evaluation metric'' (\hlc[lightgreen]{green}) or ``Annotation method'' (\hlc[yellow]{yellow}) not to be confused with reference table's ``Annotations''.
    }
    \label{fig:example-table}
\end{figure*}

Conducting literature reviews by reading and synthesizing information across a large set of documents is vital for scientists to stay abreast of their fields yet is increasingly laborious as the number of scientific publications grows exponentially \cite{jinha2010article,bornmann2021growth}. 
At the core of this sensemaking process is identifying a \emph{schema}, a set of important aspects that are useful for comparing and contrasting prior literature \cite{russell1993cost}.
The results of this process are often presented in the form of \emph{literature review tables}, whose rows are a set of papers and whose columns are a set of aspects that the papers share (Figure \ref{fig:task-format}).

In this work, we conceptualize the task of literature review table generation by decomposing it into two sub-tasks:
(1) \emph{Schema-generation}: Determining a set of relevant shared aspects given a set of input papers, and %
(2) \emph{Value-generation}: Determining the value given an aspect and a paper. 
For example, a table for a set of computer vision papers on video datasets (rows) might have a schema with aspects like ``\texttt{task}'' or ``\texttt{size}'' (columns); cell values under the ``\texttt{task}'' column may say ``\texttt{VQA}'' or ``\texttt{classification}'' (values).

Prior work has largely investigated each of the two sub-tasks independently. In particular, the large body of literature on document-grounded question-answering~\citep{kwiatkowski-etal-2019-natural,dasigi-etal-2021-dataset,qasa2023lee}, information extraction~\citep{luan-etal-2018-multi}, and query~\citep{zhong-etal-2021-qmsum,xu-lapata-2020-coarse} or aspect-based summarization~\citep{yang-etal-2023-oasum,ahuja-etal-2022-aspectnews} advances methods that are also suitable
for generating values conditioned on an aspect.
In our example above, values for aspect ``\texttt{size}'' can be answers to questions like ``How many videos are in this dataset?''. 

In contrast, schema generation from a set of documents remains relatively under-explored, even though it is a crucial and effortful part of the manual literature review process.
Prior work like \citet{Zhang2018OntheflyTG} infers new schemas from pre-existing ones, while recent work like \citet{wang2024scidasynth} assumes users can clearly articulate a schema in a short natural language query to infer aspects directly.
This paper studies the use of language models for literature review table generation with a focus on unifying these two sub-tasks. This presents us with two research challenges:

First, we note a lack of large-scale, high-quality datasets of literature review tables to serve as a benchmark for this task. 
Second, similar to challenges faced in summarization and other grounded generation tasks, semantically similar content can be expressed with different surface forms, which makes automatic evaluation difficult even with a high-quality dataset. An example of these surface form differences is in Figure~\ref{fig:example-table}.
To address these challenges:

\begin{itemize}
    \item In \S\ref{sec:data-curation}, we curate and release \ourdata,\footnote{DIGESTables stands for \textbf{D}ocument \textbf{I}nformation \textbf{G}athering and \textbf{E}xtraction for \textbf{S}cientific \textbf{T}ables} %
    a dataset of \numtotaltables{} high-quality literature review tables scraped and filtered from 16 years of ArXiv papers uploaded between April 2007 and November 2023. These tables compare and contrast a total of \numtotalpapers{} unique papers using a total of 7,634 columns and 43,905 values. This is the result of extensive filtering on an initial set of around 2.5 million extracted tables to ensure high quality, based on a strict set of desiderata. Finally, we link every table to rich paper content: (1) every input paper (row) has corresponding full text document, and (2) every table has its caption and in-line textual references extracted from the table's source paper for contextual information.
    \item In \S\ref{sec:auto_metric}, we present \ourmetric{}, an automatic evaluation framework for comparing model-generated and human-authored tables.
    Our approach overcomes the difficulty in matching semantically-similar but lexically-different column names by using a language model to expand column names into descriptions grounded in documents.
    Combining with a small textual similarity model results in a matcher that is nearly twice more precise than prompting Llama~3~(70B), which often hallucinates matches.
\end{itemize}

We formalize the literature review table generation task (\S\ref{sec:table-reconstruction}) and introduce our framework for literature review table generation and detail our implementations using open and closed models (\S\ref{sec:table-generation}).

Finally in \S\ref{sec:eval}, we evaluate LMs on this generation task, addressing two key questions: (1) what contextual information is needed to steer language models to reconstruct human-authored schemas? and (2) are generated aspects that \textit{don't} match gold still useful? For (1), we find that language models have higher recall by conditioning on more context that specifies the purpose of the table (e.g., captions, in-line references, other example tables). For (2), we find that novel aspects not in the reference tables can still be of comparable usefulness, specificity, and insightfulness. %

\section{Creating \texorpdfstring{\ourdata{}}{ArxivDIGESTables}}
\label{sec:data-curation}

\paragraph{Desiderata} To enable research in synthesizing 
literature review tables, we first collect and curate a set of reference tables to ground our task and enable evaluation. To ensure this data is realistic, high-quality, and focused on supporting literature review, we decide on the following desiderata for including tables in our \ourdata{} dataset:
\begin{enumerate}[leftmargin=*,topsep=0pt]
\setlength\itemsep{-0.1em}
    \item Tables should be ecologically valid---reflecting real syntheses authored by researchers rather than artificial annotation;
    \item Tables should be focused on summarizing multiple aspects of a set of papers as opposed to tables for reporting empirical results;
    \item Tables should follow a common structure where each row represents a single document and each column represents a specific aspect.
\end{enumerate}
Based on these goals, we used the procedure below to construct \ourdata{}:

\paragraph{Data Source}

\begin{figure*}
    \centering
    \includegraphics[width=1\linewidth]{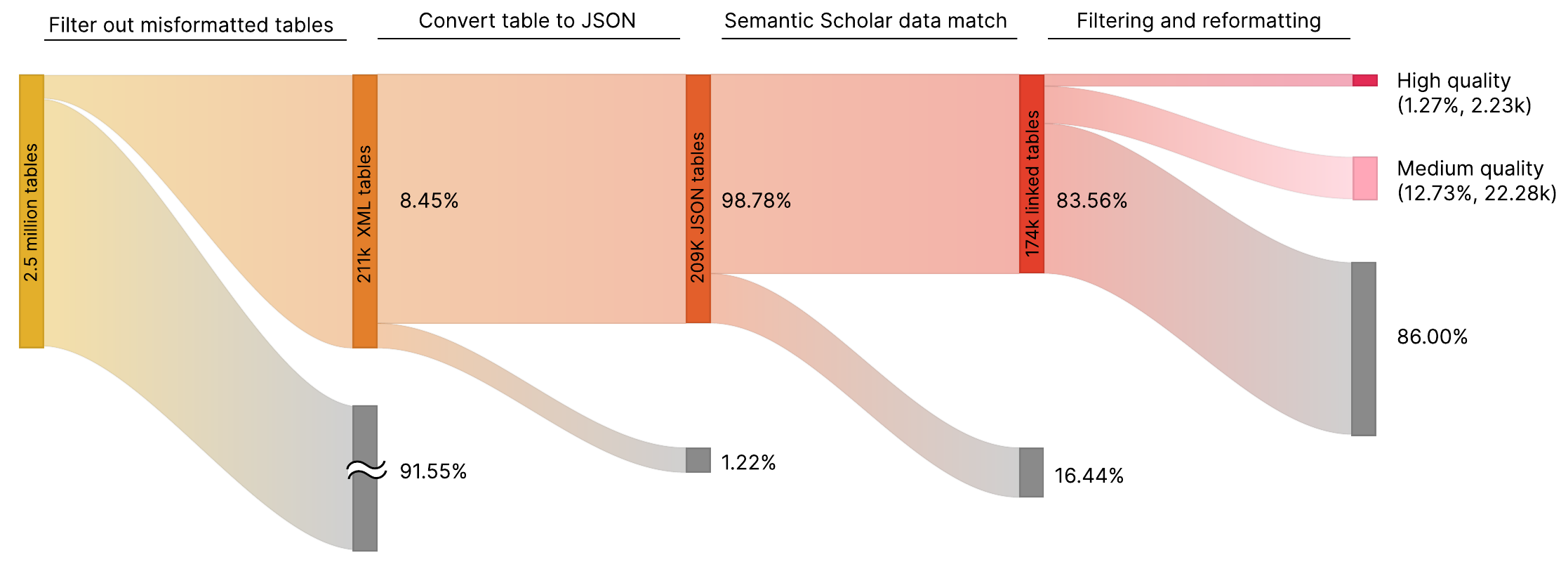}
    \caption{Pipeline for curating \ourdata{} involves extensive data cleaning and filtering. The full pipeline filters from 2.5 million starting tables published in 800,000 papers to 2,228 tables published in 1,723 papers. Data pipeline described in \S\ref{sec:data-curation}.}
    \label{fig:our-data-implementation}
\end{figure*}

To ensure our task and benchmark are grounded in realistic cases, we collected a dataset real-world literature review tables from open access ArXiv papers from April 2007 until November 2023.
We subsequently filter these tables down to a high-quality set of \numtotaltables{} tables that meet our desiderata, as seen in Figure~\ref{fig:our-data-implementation}. %

\paragraph{Extracting Tables}
The first step in our data collection pipeline is to extract the tables from papers published on the ArXiv preprint server.
To start, we consider approximately 800,000 papers that have LaTeX source available.
We then use unarXive~\citep{Saier2023unarXive} to convert the ArXiv source into XML.
From these XML documents, we extract $\sim$2.5 million tables. %

\paragraph{Filtering Tables}
As a first filtering pass, we remove tables that are likely to be misparsed or unusable, filtering those with fewer than 400 or more than 15,000 characters. We also remove tables that have no table cell tags within them.
Toward Desiderata 3, we filter out tables that have fewer than two citations, two rows, or two columns.
We also remove any tables that have citations in more than one column, as these are often tables where papers are values rather than rows. This leaves approximately 211,000 tables.

\paragraph{Matching Rows to Papers} 
We use heuristics to convert XML-formatted tables into JSON objects that allow us to directly index the tables by paper and aspect (See \S\ref{appendix:data-xml-processing} for details).
At this stage, the citation information is usually contained within a cell in a table.
For instance, an example cell with the header ``\texttt{Model}'' might have the value ``\texttt{BERT (Devlin et al., 2019)}''.
We extract the citations from these cells and place them in their own column called ``\texttt{References}''. 
Rows without citations are assumed to refer to the source paper containing the table.
After this step in the process we remove any tables where the algorithm failed and any tables that now have fewer than two rows, leaving 47,876 tables.

\paragraph{Obtaining Table Citation Metadata}
unarXive \cite{Saier2023unarXive} helpfully links each citation in the table to a bibliography item.
We use endpoints from the Semantic Scholar API~\citep{kinney2023semantic} to obtain titles and abstracts.
This occasionally fails for various reasons (e.g., the bibliography text is missing information, the paper is missing from or could not be found in the Semantic Scholar database).
We filter out any tables that have fewer than two matched citations, leaving us with 44,617 tables.

\paragraph{Grounding to Paper Texts}

To meet Desiderata 2, we want to ensure that the information in the table actually comes from the cited paper.
For instance, a common type of table reports experimental results whose values require actual experimentation and cannot be derived from the input papers' text alone.
To filter such columns, we remove any that have math symbols or floating point numbers.
Additionally, to make sure the generation task is tractable, we remove any rows whose papers do not have publicly-available full texts.

\paragraph{Final Filter and Manual Verification} The last step applies a set of stringent filters and manually identifies and corrects any parsing errors (Details in \S\ref{appendix:data-high-qual}). Finally, we produce a set of \numtotaltables{} high-quality tables. (See Appendix~\S\ref{appendix:example-instance} for a sample instance.)

\paragraph{Dataset Statistics}
We present summary statistics in Table~\ref{tab:data_hq_stats} of our high-quality set of \ourdata.\footnote{To enable future work to improve on this pipeline, we also release a set of $22,283$ medium-quality tables (see Figure~\ref{fig:our-data-implementation}) with less strict filtering alongside which filters we ran to produce it along with quality metadata (See \S\ref{appendix:data-med-qual}).%
}
We are also interested in the types of aspects represented in the tables, the topics of the columns, and the fields the tables come from.
To categorize the table aspects, we use simple heuristics (Table~\ref{tab:data_hq_aspect_dist}).
We find $\sim$40\% of the columns are categorical or boolean, which are more suitable for supporting inter-paper comparisons, while the other $\sim$60\% are more descriptive.
To obtain column topics, we manually annotate columns in $\sim$50 tables---$\sim$38\%
are about datasets, $\sim$20\%
are about methods, 
and the rest are on other topics such as applications or tasks.
Finally, we use the ArXiv API to obtain which archive a table's paper was submitted to.
We find a majority (1,985) of the tables come from computer science publications, with others coming from Physics, Quantitative Biology, Statistics, Math, and other fields (See Appendix \ref{appendix:data-fos}).

\begin{table}[h]
    \centering
    \small
    \renewcommand{\arraystretch}{0.5}
    \begin{tabular}{cccccc}
        \toprule
            & Min & Max & Median &  Mean & Total \\
        \midrule
      Papers &   1 &  35 &    3.0 & 4.944 & 11016 \\
      Aspects &   2 &  13 &    3.0 & 3.426 &  7634 \\
         \bottomrule
    \end{tabular}
    \caption{Number of papers (rows) and aspects (columns) in \ourdata{}. Of the 11,0016 total rows there are 7,542 unique papers.}
    \label{tab:data_hq_stats}
\end{table}

\begin{table}[h]
    \centering
    \small
    \renewcommand{\arraystretch}{0.5}
    \begin{tabular}{lcc}
    \toprule
    Aspect Type & \% of Cols & Example Value\\
    \midrule
    Category & 35.5\% & ``Open'' vs ``Proprietary''\\
    Entity & 27.3\% & ``CNN/Daily Mail'', ``Reddit''\\
    Numeric & 21.7\% & ``$10,000$''\\
    Text & 9.7\% & ``\dots collected via various \dots'' \\
    Boolean & 5.8\% & ``$\checkmark$'' vs ``\xmark'' \\
    \bottomrule
    \end{tabular}
    \caption{Types of aspects in \ourdata's columns.}
    \label{tab:data_hq_aspect_dist}
\end{table}

\section{Literature Review Table Generation}
\label{sec:table-reconstruction}

Equipped with our dataset, we formalize the task of generating literature review tables.

\paragraph{Task Definition} 
We define our \emph{table generation} task as follows: 
Given an input set of $M$ documents $d_1,\dots,d_M$, generate a table with $M$ rows and any number of columns $N \geq 2$. 
Each row $r_1,\dots,r_M$ corresponds to a unique input document.
Each column $c_1,\dots,c_N$ represents a unique aspect.
Taken together, the columns constitute a schema.
The table then has $N \times M$ values, with one value in each cell.\footnote{We leave the case where a cell can contain multiple values to future work.}
The cell values should be derived from the input documents.

\paragraph{Generation}

We consider two main approaches to generate a table given a set of input documents.
(1) The schema and values could be \emph{jointly} generated, e.g. in a single call to a language model. This approach is fast, but initial experiments found it more prone to hallucinations and generic column names (e.g., ``\texttt{Title}'' or ``\texttt{Year}'').
(2) The generation process can be \emph{decomposed} into separate schema and value generation steps.
This approach is slower but allows us to
overcome context window limits 
and leverage prior work in aspect-based question answering to perform value generation.

\paragraph{Evaluation}

We evaluate our approaches by determining whether the generated schemas are \emph{useful} and values are \emph{correct}.
We consider a generated schema to be useful if its aspects either match those in the corresponding human-authored table in \ourdata{} or if human evaluators rate them to be useful.\footnote{There are many alternative ways to evaluate usefulness. For example, adding constraints on users' reading time could penalize very detailed tables, while ideation-focused use cases could penalize more generic aspects.}
These two conditions allow us to measure how well systems \textit{reconstruct} reference table aspects (\S\ref{sec:calibrating-schema-alignment}) and evaluate their ability to generate \textit{novel} aspects (\S\ref{sec:human-eval-schema}).
Second, we evaluate correctness of values as we would for any information extraction or QA task: for a pair of aligned columns (and rows), we judge whether the predicted cell value is semantically equivalent to the gold cell value (see \S\ref{sec:value_eval_dec}).

\section{Experiments}
\label{sec:table-generation}
We prompt language models to perform either joint or decomposed generation.

\subsection{Base Models} We use two language models, one open-weight, Mixtral 8x22~\citep{mistral_mixtral}, and one closed weight, GPT-3.5-Turbo~\citep{chatgpt}. 
To avoid gaming our recall metric, we instruct all models to generate schemas with the same number of aspects as the corresponding reference tables. (More prompting details in Appendix~\S\ref{appendix:lm-prompting}).

\subsection{Joint Table Generation}
We represent input papers using their titles and abstracts, which usually have enough information to form useful schemas and are easier to fit in the context window of models.
We use a zero-shot table generation prompt (Appendix \S\ref{appendix:prompts-baseline}).
We treat this condition as our baseline.

\subsection{Decomposed Table Generation}

\paragraph{Step 1: Schema generation}
 Like in joint generation, we represent input papers using their titles and abstracts.
 We explore a range of prompts, each including a different piece of additional context (detailed in \S\ref{sec:types-of-context}).

\paragraph{Step 2: Value generation}
Similar to extractive QA, for each aspect-paper pair, we %
prompt the model to generate a cell value based on the aspect name and the \emph{full text} of the paper.
After generating values for each paper given an aspect, we instruct a model to rewrite the values to be shorter and more consistent in style for display in table format. For this step, we use GPT 3.5-Turbo for speed and accuracy~\citep{chatgpt} (prompt in Appendix \S\ref{appendix:prompts-value-generation}).

\subsubsection{Additional Context}
\label{sec:types-of-context}

To further investigate what contextual information is needed to steer language models to reconstruct human-authored tables, we test the following additional contexts, which could be added to either schema and/or value generation (see Appendix~\S\ref{appendix:prompts} for prompts): 
(1) a generated caption where GPT-3.5-Turbo generates a short description that is consistent with all input papers; (2) the gold caption from the reference table; (3) the gold caption and in-text references, which include referencing sentences from the table's source paper; and (4) few-shot in-context examples, consisting of five reference table examples from \ourdata{} retrieved based on cosine similarity between caption embeddings~\citep{reimers-2019-sentence-bert}.

\section{Developing an Automatic Metric}
\label{sec:auto_metric}

Below we describe the design of our automatic evaluation procedure with
two components: evaluating the schema and values for a generated table.

\subsection{Schema Evaluation}
\paragraph{Challenges} The key challenge in assessing how well a generated table reconstructs a reference table lies in \textit{determining schema alignments}---identifying which columns convey the same information despite different phrasing. Two issues make schema alignment difficult. First, reference tables tend to present information concisely, making column headers and values hard to interpret without additional context (e.g., a column might be named ``\texttt{VQA}'' instead of ``\texttt{video quality assessment}''). Second, information in generated and reference tables might have low lexical overlap despite semantic similarity, a problem also observed in summarization evaluation~\citep{rouge2004package}.

\paragraph{Problem Definition} To formalize the schema alignment problem, recall that a table schema is a set of $N$ aspects. Given a model-generated table schema, $S^m = \{a^m_1, \dots, a^m_N\}$, a reference table schema $S^r = \{ a^r_1, \dots, a^r_N\}$, and a threshold $0 \leq t \leq 1$, our goal is to construct a scoring function $f$ to score each pair of aspects, $(a^m_i, a^r_j)$, such that $f(a^m_i, a^r_j) > t$ if and only if human raters would agree that $a^m_i$ and $a^r_j$ convey the same information.

\paragraph{Alignment Framework} We propose to define $f$ as the composition of two functions: a featurizer ($\phi$), and a scorer ($g$). The goal of the featurizer is to improve aspect interpretability by incorporating additional context, while the goal of the scorer is to account for meaning-preserving lexical diversity, leading to better schema alignments. 

\paragraph{Configurations of $f$}\label{sec:configurations-of-f} We study three featurizers $\phi$: (1) ``\texttt{name}'' only takes the column name as-is, (2) ``\texttt{values}'' concatenates all values under a column to the name, and (3) ``\texttt{decontext}'' prompts a language model\footnote{\texttt{Mixtral-8x7B-Instruct-v0.1}~\citep{mistral_mixtral}.} to generate a stand-alone description~\citep{choi-etal-2021-decontextualization,newman-etal-2023-question}, given the column name and its values.

\noindent We also study four scoring functions $g$:
\begin{itemize}[leftmargin=*,topsep=0pt]
\setlength\itemsep{-0.1em}
    \item \textbf{Exact Match}, which assigns a score of $1$ if $\phi(a^m_{i}) = \phi(a^r_{j})$ and $0$ otherwise.
    \item \textbf{Jaccard}, which computes Jaccard similarity of the featurized aspects, with stopwords removed.
    \item \textbf{Sentence Transformers}, which encodes featurized aspects using \texttt{all-MiniLM-L6-v2} and computes cosine similarity between them \cite{reimers-2019-sentence-bert}.
    \item \textbf{Llama 3}, which prompts Llama 3 (70B) Chat with generated and reference tables, with the column headers replaced by featurized versions, instructions to output aligned columns, and ten in-context examples. All pairs of columns returned by the LLM are assigned a score of $1$, and $0$ otherwise. Refer
    to \S\ref{appendix:prompts-llama3-scorer} for prompting details.
\end{itemize}

\paragraph{Calibrating Schema Alignment}
\label{sec:calibrating-schema-alignment}
\begin{figure}[h]
    \centering
    \includegraphics[width = \linewidth] {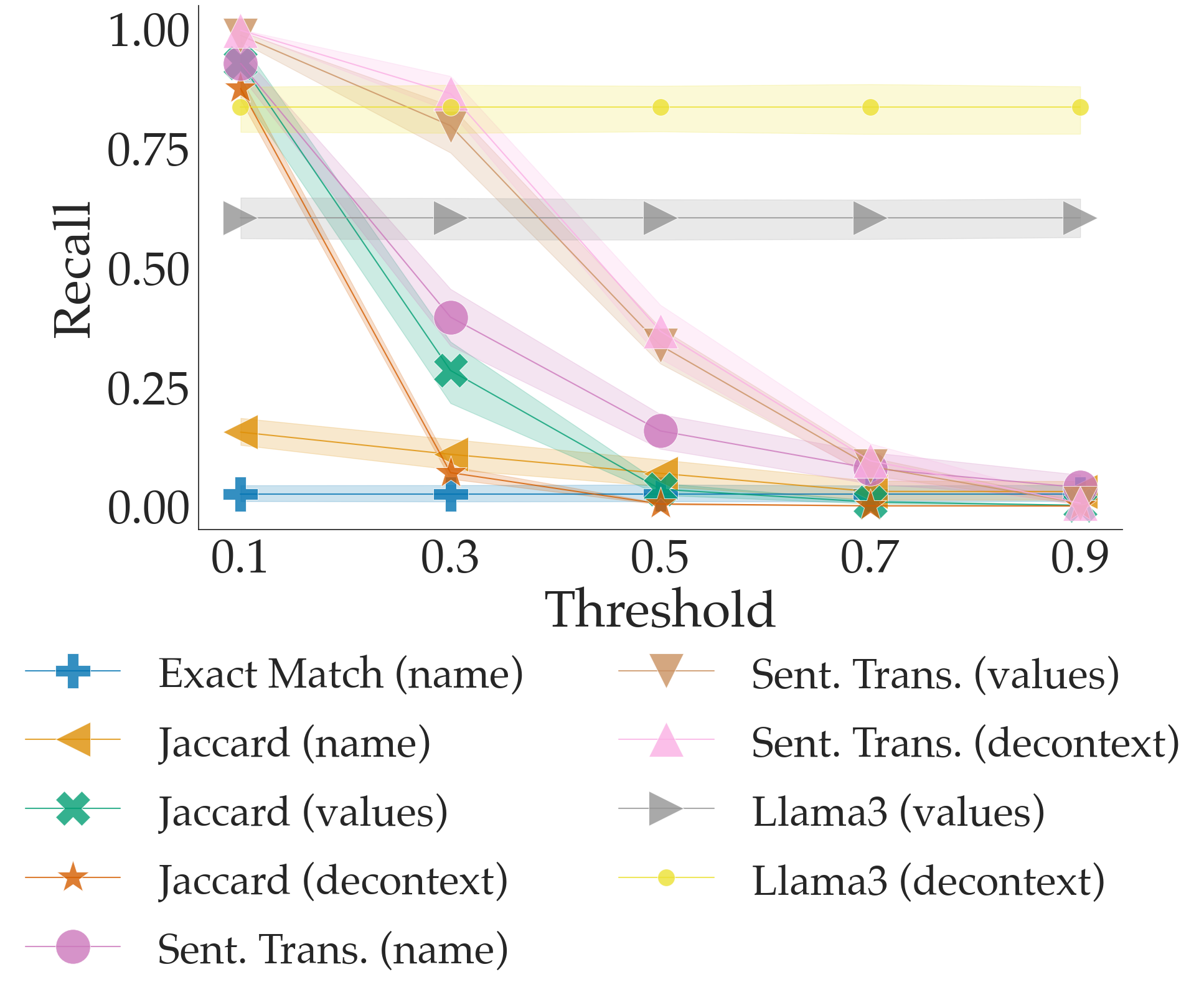}
    \caption{Recall averaged over different contexts and systems. The band represents 95\% confidence interval. Llama3 scorers have high recall, but low precision. Sentence Transformers (decontext) has the best trade-off.}
    \label{fig:metric-schema-recall}
\end{figure}

We first run various combinations of $(\phi, g, t)$ and compute schema recall (i.e., proportion of reference table aspects matched to generated table aspects) on 25\% of the tables in \ourdata{}. In Figure \ref{fig:metric-schema-recall}, we observe a wide range of recall trade-offs: 
(1) Exact match has very low recall, as expected, serving as our conservative bound.
(2) Llama 3 aligners tend to predict many more matches than other configurations despite that half of the in-context examples are tables with no matches. Llama 3 aligners serve as our upper bound. We perform human evaluation on $\sim$50 tables and find that Llama 3 aligners have between 37--55\% precision on their predicted matches.\footnote{Predicted matches are rated either as incorrect, partially, or completely correct. The lower bound only counts complete matches and the upper bound includes partial matches.} %
(3) Focusing our attention on the configurations that yield recall between these two bounds, we evaluate a range of configurations on the same tables and arrive at \ourmetric{}, our best configuration with $\phi$ using \texttt{decontext} features, $g$ using sentence transformers, and $t= 0.7$; we find \ourmetric{} performs at 70--85\% precision with acceptable yield. %

\subsection{Value Evaluation}
\label{sec:value_eval_dec}
Automated value evaluation suffers from the same issues that complicate schema evaluation, but one issue specific to value evaluation is reliance on \emph{accurate schema alignments}. If aspects are incorrectly matched %
by a schema alignment metric, performance on value evaluation might rise/drop undeservedly. Therefore, we propose evaluating value generation in isolation, instead of an end-to-end table evaluation setting. 

Specifically, we use the reference table's schemas as input to our value generation module. This ensures that every value in the reference table has a corresponding generated value (barring generation failures), bypassing the need for schema alignment. Following \S\ref{sec:types-of-context}, we consider three settings using different types of contexts: (1) ``\texttt{Column Names}'' only, (2) ``\texttt{Caption Context}'' which adds the table caption, and (3) ``\texttt{All Context}'' which further adds in-text references. Prompts used for each setting are in Appendix \S\ref{appendix:prompts-value-generation}. We then use the same suite of scorers from \S\ref{sec:configurations-of-f} (except Llama 3, which we observed was low-precision) to compute overlap between pairs of generated and reference table's values.

\section{Results}
\label{sec:eval}

\subsection{Schema Evaluation Results}
\label{ssec:schema-eval}
\begin{figure}
    \centering
    \includegraphics[width=\linewidth]{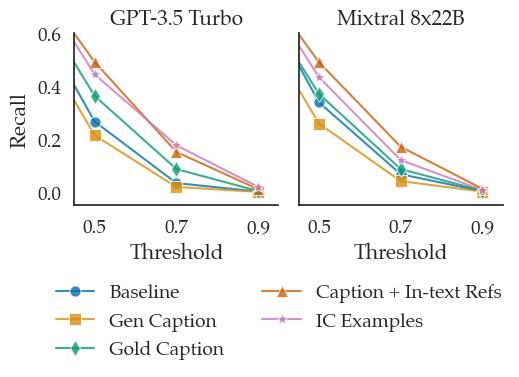}
    \caption{Schema recall for GPT-3.5-Turbo and Mixtral 8x22, using various types of additional contexts. All scores are computed using our best metric: sentence transformer-based scorer with decontext featurizer. More context improves recall, but does not lead to completely reproducing reference table schemas.}
    \label{fig:schema_results}
\end{figure}

\paragraph{Automated Evaluation} Figure~\ref{fig:schema_results} shows the ability of GPT-3.5-Turbo and Mixtral 8x22 to reconstruct schemas (as measured via \ourmetric) using various types of additional contexts described in \S\ref{sec:types-of-context}. 
Turning back to the question: \emph{How does the amount of context provided affect table reconstruction?}
(1) We see that \textbf{low context} prompts (e.g., a baseline with no additional context, caption-only) perform the worst while \textbf{high context} prompts (e.g., in-text references, in-context examples) perform best. This trend is fairly stable across systems. 
(2) Interestingly, though adding context improves reconstruction, it does not make the task trivial --- even the best performing systems are far from perfect.

One potential concern for this analysis is that the models we use may have seen the older tables during training, which could inflate performance.
To address this, we compute recall separately on subsets of newer and older tables (those from before or after January 2023 constituting 30\% and 70\% of our data respectively) for the high context prompts.
We find that there is minimal difference between these two sets (the newer tables have recalls on average 1--3 percentage points lower).

\paragraph{Human Evaluation} 
\label{sec:human-eval-schema}
Our automated evaluation measures how well LMs can recover the reference tables' aspects, but leaves an additional question: \emph{Are LM-generated novel aspects which do not match with gold aspects also useful?} To investigate this, we collect human assessments of generated aspects. Annotators are provided a generated table and the titles of all input papers. They are then prompted to provide a 5-point Likert scale rating for each of the following aspects: (1) general \emph{usefulness} for understanding the input papers, (2) \emph{specificity} to input papers (i.e., would this aspect be applicable to any other set of papers), and (3) \emph{insightfulness} of the generated aspect (i.e., capturing novelty).\footnote{See Appendix \S\ref{appendix:human-eval} for the annotation interface and the definitions used.} We also instruct annotators to only judge based on the quality of the \emph{aspects} only, ignoring the values which are evaluated separately. After collecting these ratings, we separated the rated aspects into two groups---ones that matched a gold aspect (\textbf{M}), and ones that did not (\textbf{NM}). The annotators were blind to the conditions when rating the aspects, and inter-annotator agreement was $0.56$ (Krippendorff’s $\alpha$). %

Comparing ratings on matched and unmatched aspects, we did not find aspects that matched to be rated significantly higher than ones that did not (Table~\ref{tab:schema-eval-human}; Mann-Whitney U tests). This suggests that novel generated aspects are of comparable quality (\emph{usefulness}, \emph{specificity}, \emph{insightfulness}) to gold aspects or even have a higher quality (\emph{usefulness} of aspects from Caption+In-text References). Moreover, aspects from Caption+In-text Reference are shown to be more \emph{useful} and \emph{specific} than the Baseline's, but were less \emph{insightful}.
This suggests an interesting tradeoff between our reconstruction objective, and possibly a different objective like creativity.

\begin{table}
    \centering
    \small
    \renewcommand*{\arraystretch}{1.2}
\begin{tabular}{@{\hskip .1em}l@{\hskip .5em}c@{\hskip .5em}c@{\hskip .5em}c@{\hskip .5em}c@{\hskip .5em}c@{\hskip .1em}}
    \toprule
& \multicolumn{2}{c}{\textbf{Caption+In-text Ref}} & \multicolumn{2}{c}{\textbf{Baseline}}  & \\
& \textbf{M} & \textbf{NM} & \textbf{M} & \textbf{NM}  \\
    \midrule
\textbf{Useful} & 3.70 (1.74) & 4.07 (1.06) & 3.92 (0.69) & 3.73 (1.17) \\
\textbf{Specific} & 2.88 (1.26) & 3.06 (1.34) & 2.85 (1.31) & 2.75 (1.35) \\
\textbf{Insightful} & 1.86 (1.04) & 1.93 (1.21) & 2.34 (1.25) & 2.27 (1.19) \\
    \midrule
\textbf{\# Samples} & 102 & 208 & 64 & 283\\
    \bottomrule
    \end{tabular}
    \caption{Mean (SD) ratings from human assessments of generated aspects that match the gold schema (M) with those that do not (NM).
    }
    \label{tab:schema-eval-human}
\end{table}

\paragraph{Error Analysis}
Finally, we report some qualitative observations of errors in the generated schemas we used for human evaluation.
These point to future areas of improvement.
Comparing outputs from the baseline to the Caption+In-text References condition, we find that the latter tends to output more specific aspects. For example, for one table, the Mixtral baseline produces aspects ``Model Architecture'' and ``Application'', while the Caption+In-text References Mixtral system generates the more specific aspects ``Maximum resolution'' and ``Training batch size''.
We also note a few differences between schemas generated in the Caption+In-text references setting the reference tables' schemas, as well as categories of aspects that can pose difficulty for generation in Table~\ref{tab:errors-schema} and additional examples in Appendix~\S\ref{appendix:qual-error-analysis}.

\subsection{Value Evaluation Results}
\begin{figure}
    \centering
    \includegraphics[width = \linewidth]{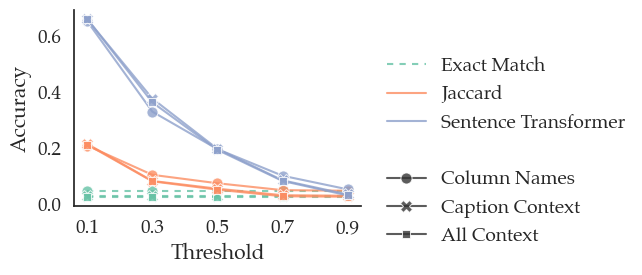}
    \caption{Value generation accuracy for GPT-3.5-Turbo using various types of additional contexts, as computed by different scorers.}
    \label{fig:metric-value-recall}
\end{figure}

\paragraph{Automated Evaluation.} Figure~\ref{fig:metric-value-recall} shows the performance of GPT-3.5-Turbo on value generation, using various types of additional contexts (described in \S\ref{sec:value_eval_dec}.) We see that scorers continue to follow the same trend observed during schema alignment, with the sentence transformer scorer being fairly permissive while an exact match is overly strict. Interestingly, unlike schema reconstruction, we observe that incorporating additional context does not seem to improve value generation accuracy; we dig deeper into this during human evaluation. Finally, like schema alignment, models are far from perfect in value generation.

\paragraph{Human Evaluation.} We conduct additional human evaluation to investigate whether adding context indeed has no impact on value accuracy, or our automated metrics are not sensitive enough to capture differences. We randomly sample 30 tables and compare gold vs generated values for these tables under all three settings. For each gold-generated value pair, we have two annotators label whether it is a complete match, partial match or unmatched. Partial matches include cases where values are lists of items and the generated value misses or adds some (e.g., ``DPO'' vs ``DPO, PPO''), or cases where the gold and generated values have a hypernymy relationship (e.g., ``graph neural networks'' vs ``GATs'').
Inter-annotator agreement is $0.55$ (Cohen's $\kappa$).
Table~\ref{tab:value-eval-human} presents results from this assessment, showing that adding additional context leads to a significant improvement in partial matches. However, many matches have no lexical overlap (e.g., ``X'' vs ``No'') or require some inference (e.g., ``Yes'' under a column called ``sensors deployed'' should match a value like ``sensors used to monitor air quality'').
This indicates that there is scope for further research in developing more sensitive featurizers and scorers for value evaluation.

\begin{table}
    \centering
    \small
    \renewcommand*{\arraystretch}{1.2}
    \begin{tabular}{@{\hskip .1em}l@{\hskip .5em}c@{\hskip .5em}c@{\hskip .5em}c@{\hskip .1em}}
    \toprule
\textbf{Setting} &  \textbf{Complete} & \textbf{Partial}  & \textbf{None} \\
    \midrule
\textbf{Col. Names} & 21.13\% (75) & 22.54\% (80) & 56.34\% (200)\\
\textbf{+ Captions} & 18.84\% (65) & 31.30\% (108) & 49.86\% (172) \\
\textbf{+ IT-Refs} &  22.65\% (77) & 31.77\% (108) & 45.59\% (155) \\
    \bottomrule
    \end{tabular}
    \caption{Proportion of matched gold-generated value pairs for various context settings, according to human assessment.}
    \label{tab:value-eval-human}
\end{table}

\begin{table*}[t]
    \centering
    \small
    \begin{tabular}{p{10em}p{36em}}
        \toprule
        Challenge Type & Description   \\
        \midrule
        Different Granularity & The generated schema might be a high-level category (e.g. ``data types''), while the reference schema includes more specific aspects (e.g. ``image'', ``text'', ``audio'', etc.) \\
        Different topics & The generated schema might have a different variety of topics than the reference schema (e.g. \{``model architecture'', ``dataset used'', ``performance metric''\} versus just dataset properties \{``color'', ``context''\}) \\
        Complex Aspects & Aspects combine information from multiple cells, which can mislead the value generator. E.g. ``dataset size'' leads to some values pertaining to training data and others to test data. \\ 
        Overly Specific & A predicted aspect might only apply to one paper \\
         \bottomrule
    \end{tabular}
    \caption{Qualitative observations of challenges with generated tables}
    \label{tab:errors-schema}
\end{table*}

\section{Related Work}

\subsection{Schema Generation for Literature Review}

Synthesizing schemas from research papers has been previously studied in contexts like
identifying relations between papers \cite{shahaf2012metro,lee2024paperweaver}, organizing research threads \citep{kang2023synergi}, discovering papers for ideation \citep{hope2022scaling, kang2022augmenting}, or constructing intermediate scaffolds for better multi-document summarization ~\cite{ Shah2021NutribulletsHC}. These works often assume fixed or sparse schemas, focus on a sub-%
component of schema generation, or do not evaluate intermediate tables. More closely related to our work, SciDaSynth is an interactive interface for creating ``data tables'' from a set of papers~\cite{wang2024scidasynth}, which infers aspects from users' questions about the papers. However, identifying and articulating good comparison aspects can be nontrivial for users, motivating our aim of automatically inducing salient aspects. \citet{hashimoto2017automaticGO} explore automated aspect extraction for literature review tables and point out that more specific aspects are useful but hard to generate.%

\subsection{Datasets for Scientific Table Generation}
Prior work has also released datasets of tables %
~\cite{Bai2023SchemaDrivenIE,gupta-etal-2023-discomat}. \citet{Bai2023SchemaDrivenIE} build a dataset of numeric result tables, while \citet{gupta-etal-2023-discomat} release 4.4k distantly supervised and 1.5k manually annotated tables with material compositions from papers. Unlike ~\ourdata{}, these datasets do not necessarily link tables to input papers.
Multi-document summarization datasets, like Multi-XScience~\citep{lu-etal-2020-multi-xscience} and MS\^{}2~\citep{deyoung-etal-2021-ms}, are related to table generation but yield sparse tables or use fixed schemas. Finally, there are datasets for other table-related tasks such as table extraction from PDFs ~\cite{gemelli2023cte}, table retrieval~\cite{gao2017scientific}, column annotation~\cite{korini2022sotab}, table-to-text generation~\cite{Moosavi2021SciGenAD}, table transformation~\cite{chen2020wikitablet}, and table generation ~\cite{wu2021text}. 
However, these datasets either do not focus on scientific tasks or comparing papers. %

\subsection{Automated evaluation using LMs}
As LMs have improved, they have also increasingly been used for automatic evaluation across NLP tasks, including summarization and QA that our work is similar to~\cite{Zhang2020BERTScore,wang2023chatgpt, lu2024llmscore,Zheng-2024-LLM-as-a-judge,murahari2023qualeval}. 
Some work on table generation has used a combination of automated and human evaluation. \citet{hashimoto2017automaticGO} use ROUGE~\citep{rouge2004package} and human evaluation~\citep{nenkova2007pyramid} to evaluate generated summaries of a table.~\citet{Zhang2018OntheflyTG} evaluates schema selection via automatic entity ranking using ground truth entities. These works largely focus on measuring content overlap, whereas our automated metric incorporates table structure and context and our human evaluation focuses on downstream utility.

\section{Conclusion}
Language models have the potential to help scientists organize papers during literature review by synthesizing tables with schemas that aid comparison.
In this work, we curate \ourdata{}, a dataset of such tables and additional contexts that can be used to evaluate systems' abilities to produce such tables.
We present \ourmetric{}, an automatic evaluation framework for comparing model-generated and human-authored reference tables.
We then use this evaluation framework to investigate two research questions: what context is needed to reconstruct human-authored tables, and whether generated aspects that don't align with references are also useful, specific and insightful.
We release our artifacts to help spur development of literature review table generation systems, and seed potential for their role in evaluating systems' scientific synthesis abilities.

\section*{Limitations}

\paragraph{We only study scientific papers from ArXiv.} While in theory, scientists in many fields produce literature review tables, we restrict our reference tables to ones that we can scrape from ArXiv.
This means many of the papers in our dataset come from fields that are most represented on ArXiv (e.g. computer science) and fewer come from medicine, humanities, or social science publications. Additionally, all of the tables in our high quality set are in English, even though literature review tables may also be used in other languages.

\paragraph{Reconstructing tables is difficult.}
While \ourmetric{} is effective at matching generated and reference table columns, and we test providing different additional context to steer the table generation models, many generated table columns do not match with the reference columns.
Though we presented a human evaluation protocol that showed utility for generated columns that do not match the reference columns, such evaluation is costly.
Future work should investigate automatic metrics that correlate with human utility evaluations as well.

\section*{Ethical Considerations and Broader Impact}

\paragraph{Generated literature review tables might misrepresent authors' work.}
Generating literature review tables requires taking aspects of papers out of their original context to show them to users.
Similar to summarization, this process has the potential to misrepresent the original work either due to the table cell values not having enough context, or less accurate models introducing hallucinations.
Additional checks would have to be implemented if such tables were to be deployed in user-facing situations.

\paragraph{Literature review tables may discourage reading original sources.}
The resource we present is meant to encourage the development of methods to construct literature review tables.
If the field iterates on this task and develops systems that perform very well, the tables may have all of the information that a given reader wants to see.
This could discourage readers from finding the original source of the claims.
That said, the rows in the tables in our benchmark do include citations, so readers can trace values back to their sources.
However, readers are not guaranteed to follow these citations, so generated tables could encourage poor scholarly practices.

\bibliography{references}
\appendix
\lstdefinestyle{mystyle}{escapechar=@}
\lstset{
  basicstyle=\ttfamily,  %
  literate={``}{{\textquotedblleft}}2 %
   {''}{{\textquotedblright}}2 %
   {'}{{\textquotesingle}}1   %
   {`}{{\textasciigrave}}1    %
}

\section{Data Processing}
\subsection{XML Parsing}
\label{appendix:data-xml-processing}
At this point in the pipeline, the tables we are considering are represented in XML format.
Unfortunately, sometimes XML-formatted tables have column headers that span multiple rows, rows can have insufficient numbers of columns, cells may span multiple columns rows, etc.
This makes it hard to enforce Desiderata 3.
To address these difficulties, we design heuristics to parse the XML formatted tables into a JSON object that allows us to directly index the tables by paper and aspect. Our heuristics cannot be completed for all tables---sometimes they fail completely, and other times they fail on particular rows. We also experimented using GPT-4~\citep{Achiam2023GPT4TR} for these difficult cases, but still found errors due to insufficient layout information being maintained in the conversion from LaTeX source to XML.

\subsection{High Quality Data Filters}
\label{appendix:data-high-qual}
To achieve our set of tables, we apply a number of stringent filters.
We remove any tables whose headers came from merging two rows or that have a row without a citation to avoid misformatted tables.
We also deduplicate the tables using an exact string match on all columns minus the references column, and deduplicate individual rows (which includes citations).
To meet Desiderata 2, and avoid filtering out tables that have empirical results, we filter out any columns that have floating point numbers, formulas, or figures.
After these steps, we remove any tables that have fewer than two citations, rows or columns, leaving us with our final set.

\subsection{Medium Quality Data Filters}
\label{appendix:data-med-qual}
In addition to our high-quality dataset that is likely to meet our desiderata, we also release a larger set of 22,283 tables with fewer filters.
These tables are not manually checked, are filtered less stringently, and do not have linked full-texts.
In particular:
\begin{itemize}
    \item Papers in rows are required to have titles and abstracts, but not required to have full-texts. This potentially makes value generation difficult because all of the values have to come from the title and abstract.
    \item Tables are not required to have in-text references. This potentially makes schema generation difficult, as any additional context has to come from the caption (if present).
    \item Tables with at most one row with no citation are allowed, as opposed to all rows having citations.
    \item Tables with multi-row or hierarchical headers are allowed. These can sometimes lead to misformatted tables.
\end{itemize}

\subsection{Field of Study}
\label{appendix:data-fos}
A full break-down of the fields of study represented in the high-quality dataset is in Table \ref{tab:appendix-data-fos}.

\begin{table}
    \centering
    \small
    \renewcommand{\arraystretch}{0.5}
    \begin{tabular}{cc}
        \toprule
        Field & Count \\
        \midrule
        Computer Science  & 1985 \\
        Electrical Engineering and Systems Science & 131 \\
        Physics & 51\\
        Quantitative Biology & 24\\
        Statistics & 19\\
        Math & 14\\
        Quantitative Finance & 3\\
        Economics & 1\\
    \bottomrule
    \end{tabular}
    \caption{Fields of study represented in the high-quality dataset.}
    \label{tab:appendix-data-fos}
\end{table}

\subsection{Example Data Instance}
\label{appendix:example-instance}
Below is an example instance from \ourdata{}. (Some of the keys rephrased and values are elided for clarity)

\begin{lstlisting}[breaklines=true, style=mystyle]
{
Table ID: 53648c28-a2b2-4e41-...
Paper ID: 2305.14525v1
Caption: "A categorization of scope regarding design variations observed in collected corpora. The three columns are high-level design variation types, low-level details assumptions over visual designs..."
In-Text References: [
    {Section: Design Variations
    Text: "In addition to chart type, we have also observed scope...in Table {{table:<table id>}}..."}, ...
],
Table: {
    References: ["{{cite:9a81b16}}", "{{cite:d5b4bb4}}", "{{cite:342c0c4}}", "{{cite:6697498}}"],
    "Design Variation Type": ["composite arrangement", "mark and glyph", "mark and glyph", "coordinate space"],
    "Assumption": ["only multiple-view charts", "only proportion-related charts", "only timeline-related infographics", "in Cartesian coordinate space"]
},
Citation Info: [
    {
Cite ID: 9a81b16,
Title: "Composition and Configuration Patterns in Multiple-View Visualizations",
Abstract: "Multiple-view visualization (MV) is a layout design technique...",
Full Text: "1 Introduction We present an in-depth study on how multiple views are used in practice, and integrate our results into a recommendation system for the layout design..."
}, ...
}
\end{lstlisting}

\section{Prompts}
\label{appendix:prompts}

\subsection{Prompt for table generation (Baseline)}
\label{appendix:prompts-baseline}
\begin{lstlisting}[breaklines=true, style=mystyle]
System Prompt: You are an intelligent and precise assistant that can understand the contents of research papers. You are knowledgable on different fields and domains of science, in particular computer science. You are able to interpret research papers, create questions and answers, and compare multiple papers.

User Prompt: [System] 

We would like you to build a table that has each paper as a row and, as each column, a dimension that compares between the papers. You will be given multiple papers labeled Paper 1, 2, and so on. You will be provided with the title and content of each paper. Please create a table that compares and contrasts the given papers. Make {col_num} dimensions which are phrases that can compare multiple papers, so that the table has {col_num} columns. The table should also have {paper_num} papers as rows. Return a JSON object of the following format:

```json
{json_format}
```
**Check that the table has {paper_num} papers as rows and {column_num} dimensions as columns.**.

[Paper Content]
{paper1} {paper2} ... {paperN}
\end{lstlisting}

\subsection{Prompt for schema generation}
System prompt is the same as the one from table generation.
\subsubsection{Schema generation with generated captions}
\begin{lstlisting}[breaklines=true, style=mystyle]
User Prompt: [System] 

Imagine the following scenario: A user is making a table for a scholarly paper that contains information about multiple papers and compares these papers. To compare and contrast the papers, the user provides the title and content of each paper. Your task is the following: Given a list of papers, you should find aspects that are shared by the given research papers. Then, within each aspect, you should identify {num_columns} attributes that can be used to compare the given papers.

First, you should return the list of similar aspects as a Python list as follows: "["<similar aspect that all given papers shared>", ...]". Then, think of each aspect as the topic for the Related Work section of the user's paper. Finally, find attributes that can compare the given papers within the Related Work section. Return a JSON object in the following format:

```json
{{
  "<attribute 1>": ["<comparable attribute within the aspect 1>", "<comparable attribute within the aspect 1>", ...],
  ...
}}
```

[Paper Content]
{paper1} {paper2} ... {paperN}

Please ensure that your response strictly follows the given format. Adherence to the specified structure is mandatory.
\end{lstlisting}

\subsubsection{Schema generation with caption and in-text references}
Generation of schemas with captions does not include the in-text references part in the prompt below. This prompt is when the number of in-text references is $K$
\begin{lstlisting}[breaklines=true, style=mystyle]
User Prompt: [System] 

Imagine the following scenario: A user is making a table for a scholarly paper that contains information about multiple papers and compares these papers. To compare and contrast the papers, the user provides the title and content of each paper. To help you build the table, the user provides a caption of this table, which is referred to in the paper as additional information.

[Caption]
{caption}

[In-text reference]
{section header 1: in-text reference 1}{section header 2: in-text reference 2}...{section header K: in-text reference K}

Your task is the following: Given a list of papers and table caption, you should identify {num_columns} table columns to compare given research papers. Return a list in the following format:

```List
["<comparable attribute within the table caption>", "<comparable attribute within the table caption>", ...]
```

[Paper Content]
{paper1} {paper2} ... {paperN}

Please ensure that your response strictly follows the given format. Adherence to the specified structure is mandatory.
\end{lstlisting}

\subsubsection{Schema generation with few-shot examples}
\begin{lstlisting}[breaklines=true, style=mystyle]
User Prompt: [System] 

Imagine the following scenario: A user is making a table for a scholarly paper that contains information about multiple papers and compares these papers. To compare and contrast the papers, the user provides the title and content of each paper. To help you build the table, the user provides similar tables that you can refer to as follows:

{Table 1: few-shot example table 1}{Table 2: few-shot example table 2}...{Table 5: few-shot example table 5}

Your task is the following: Given a list of papers and table examples, you should identify {num_columns} table columns to compare given research papers. Return a list in the following format:

[List]
["<comparable attribute>", "<comparable attribute>", ...]
[List]

{paper1} {paper2} ... {paperN}

Please ensure that your response strictly follows the given format. Adherence to the specified structure is mandatory.
\end{lstlisting}

\subsection{Prompt for value generation}
\label{appendix:prompts-value-generation}
\begin{lstlisting}[breaklines=true, style=mystyle]
Answer a question using the provided scientific paper.

Your response should be a JSON object with the following fields:

- answer: The answer to the question. The answer should use concise language, but be comprehensive. Only provide answers that are objectively supported by the text in paper.

- excerpts: A list of one or more *EXACT* text spans extracted from the paper that support the answer. Return between at most ten spans, and no more that 800 words. Make sure to cover all aspects of the answer above.

If there is no answer, return an empty dictionary, i.e., '{}'.

Paper:
 { full_text }

Given the information above, please answer the question: "{ question }".
\end{lstlisting}

Using this strategy to generate values for columns requires the creation of questions describing the corresponding columns, for which we follow a two-step generation process. First, we prompt an LLM, specifically GPT-4-Turbo to generate descriptions for every column conditioned on additional context (either reference captions, or reference captions and in-text references). For the setting that does not use any additional context, this step is skipped.
\begin{lstlisting}[breaklines=true, style=mystyle]
CAPTION_PROMPT = """
A user is making a table for a scholarly paper that contains information about multiple papers and compares these papers. 
This table contains a column called {column}. Please write a  brief definition for this column.

Here is the caption for the table: {caption}.

Definition: 
"""

CAPTION_WITH_REF_PROMPT = """
A user is making a table for a scholarly paper that contains information about multiple papers and compares these papers. 
This table contains a column called {column}. Please write a  brief definition for this column.

Here is the caption for the table: {caption}.

Following is some additional information about this table: {in_text_ref}.

Definition: 
"""
\end{lstlisting}

Then, LLMs are prompted to rewrite generated definitions as concise queries. For the no-context setting, we use a simple template to produce queries containing the column name.

\begin{lstlisting}[breaklines=true, style=mystyle]
CONTEXT_QUERY = "Rewrite this description as a one-line question."

NO_CONTEXT_QUERY = "From the provided paper full-text, can you extract {column}?"
\end{lstlisting}

Our preliminary experiments show that the value generation module often returns empty values (in $\sim 30\%$ cases on average), which motivates us to add a retry policy. Under this policy, we generate four additional queries with minor rephrasing and retry value generation with them. We observe that this reduces the proportion of empty values to $\sim 7.5\%$. If all retries produce empty values, we return an empty value.

\begin{lstlisting}[breaklines=true, style=mystyle]
CONTEXT_RETRY_QUERIES
original_query + "Return a summary of this information"
original_query + "Try to extract this information."
original_query + "Summarize information about this."
original_query + "What information can you find about this?"

NO_CONTEXT_RETRY _QUERIES
Extract information about {column} aspect from this paper.
What information can you find about {column}?
We want to create a table comparing papers. Extract the information from this paper that goes in the column called {column}.
In a literature review table comparing multiple papers, what information from this paper would go under column {column}?
\end{lstlisting}

\subsection{Prompt for Llama-3 Scorer for Automatic Evaluation}
\label{appendix:prompts-llama3-scorer}

\begin{lstlisting}[breaklines=true, style=mystyle]
Given two tables, match column headers if their columns have very similar values. Most columns will not have a match.

Respond with a json list, whose elements are two element lists. The first element is the key of Object 1 and the matching key of Object 2.
For example, if the key 'Dataset size' and 'Number of training examples' are matched, you should return '[['Dataset size', 'Number of training examples']]. If no keys contain the same information, then just output an empty list '[]'

Table 1:
[In-context example human-authored table]

Table 2:
[In-context example generated table]

Response: [In-context example human-aligned aspects]
\end{lstlisting}

\subsection{LM Prompting Details}
\label{appendix:lm-prompting}

\begin{figure*}
    \centering
    \includegraphics[width=\linewidth]{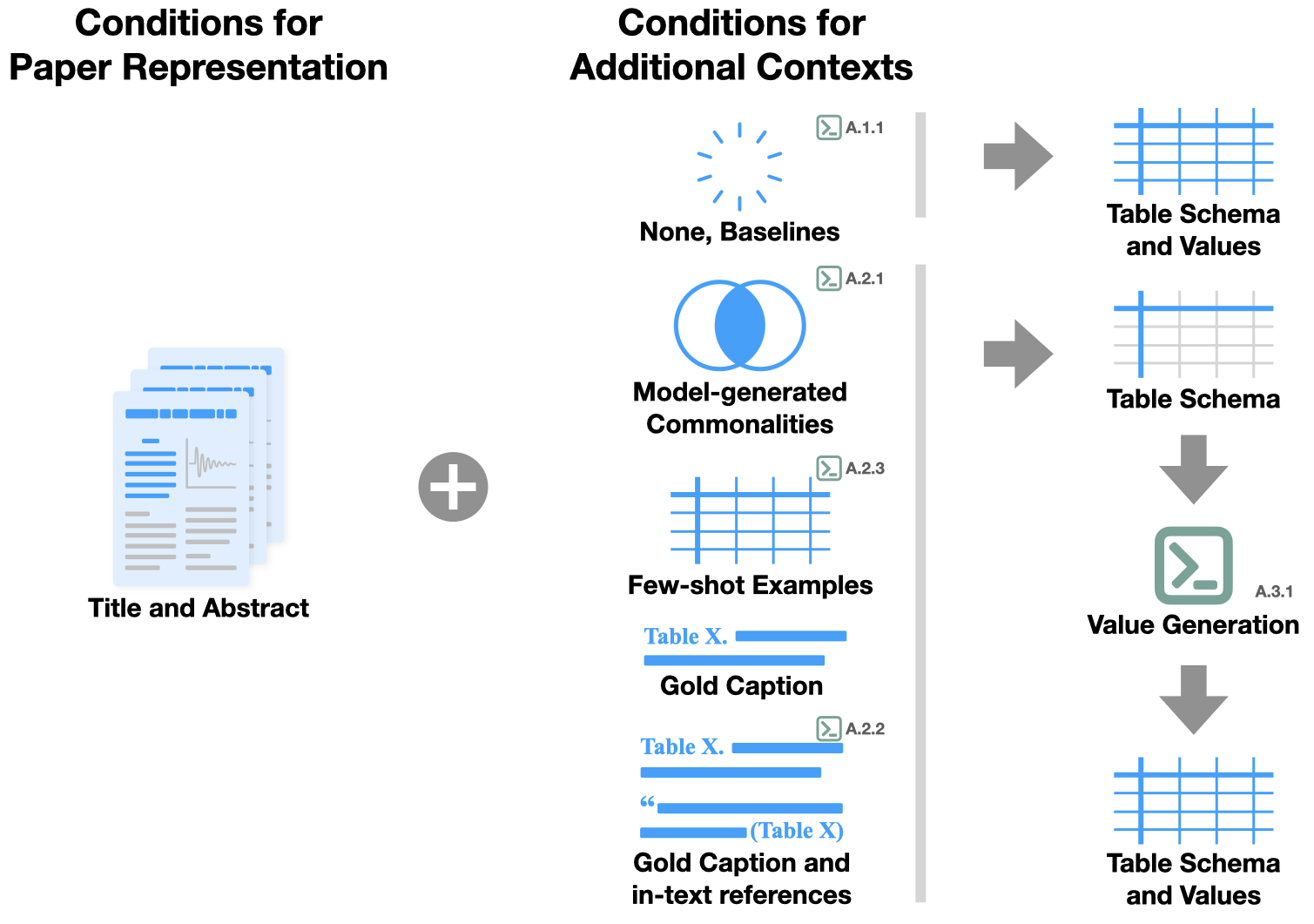}
    \caption{Diagram of prompting methods under experiment conditions.} 
    \label{fig:our-method-implementation}
\end{figure*}

\paragraph{Truncation and Error Handling.}
As our evaluation tests language models' capabilities of schema rediscovery, we implemented strategies for handling other types of errors from language model generation (e.g., a different number number of schemas between generated and reference tables, or the format of the generated output not matching with a format given in the prompt). We take both preventative as well as fall-back measures to deal with these errors:

\begin{enumerate}
    \item \textbf{Preventative}: 
    To address the issue of generating tables when context window might be insufficient due to the large number of input papers, we adopted the following approach by: (1) dividing the paper sets into smaller batches to ensure the total length of input papers does not exceed the context window size, (2) dividing the columns that need to be created into smaller batches to ensure the total number of columns from whole batches does not exceed the number of columns in human-authored tables, and (3) subsequently joining these smaller tables together without the need for further generation. The batch size is chosen based on the model and input paper representation.
    In general, though, we used a threshold of 20 abstracts per batch, determined by using the average length of the top 20\% longest abstracts to ensure that even long abstracts could fit within the context along with the in-context examples and prompts.
    We also set the number of max tokens as high as the model can handle.
    \item \textbf{Fall-back}: When encountering an error, we retry querying the model with the same prompt, and due to stochasticity in the generation process, models occasionally recover. The errors we handled with fallback strategies are as follows: (1) when the output doesn't align with the format specified in the prompt, (2) when the number of schemas, papers, and values don't match the reference table, and (3) when the entire context exceeds the context window of the base model. 
    \item \textbf{Removal}: We allow up to five retries before abandoning the input.
\end{enumerate}

\section{Human Evaluation}
\label{appendix:human-eval}
When performing human evaluation for the novel schema, we assessed each column based on the following criteria:
\begin{itemize}
    \item Usefulness:  the degree to which this column helps in understanding and comparing the set of input papers.
    \item Specificity: the degree to which a column is specific to the particular set of input papers, rather than applying to any generic set of papers.
    \item Insightfulness: the degree to which a column is about novel and deep aspects. An insightful column goes beyond surface-level information and captures novel or unexpected aspects (e.g., “\textit{Method}” column may be useful, but it may not be considered highly insightful.)
\end{itemize}

\begin{figure*}
    \centering
    \includegraphics[width=\linewidth]{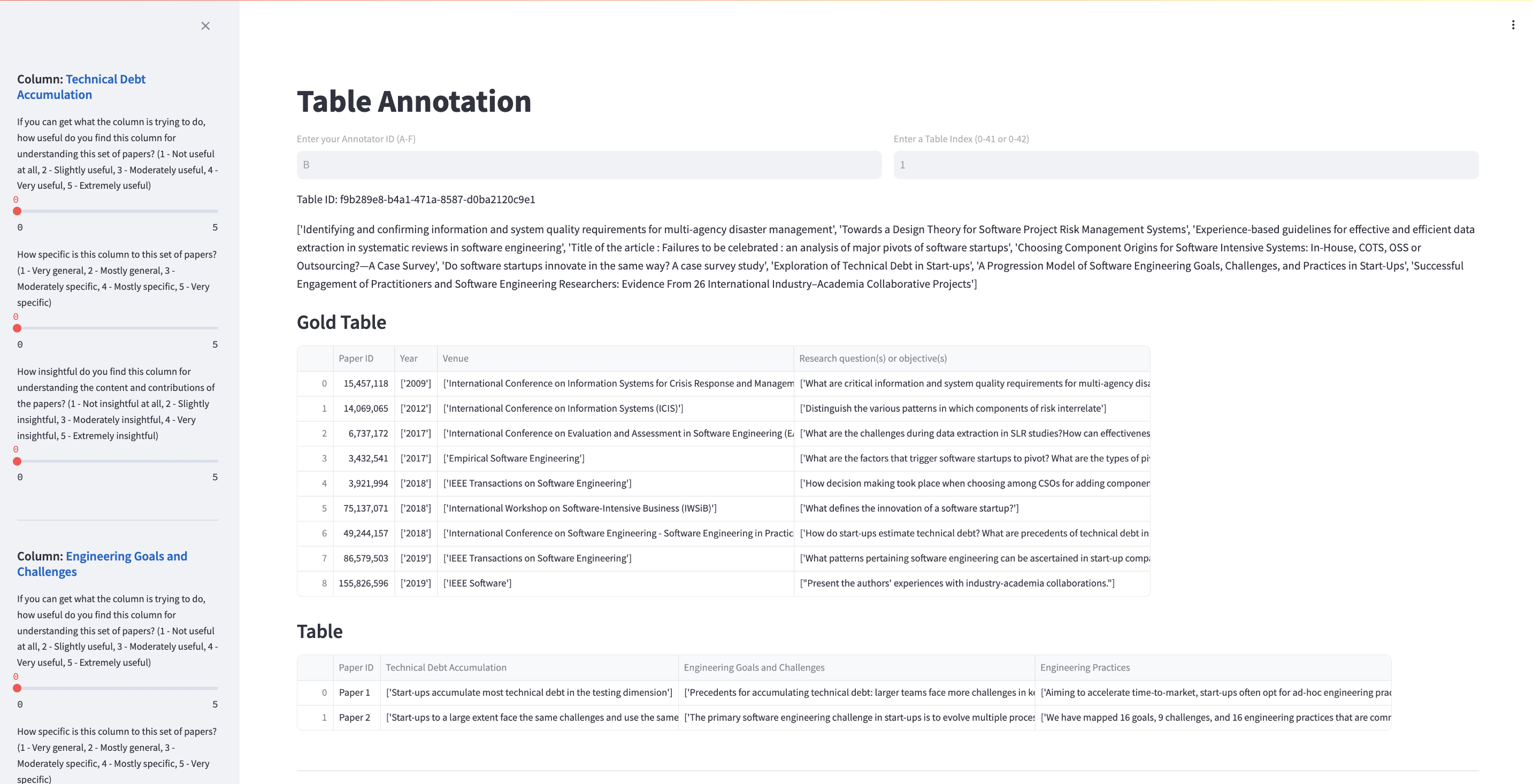}
    \caption{The interface used for annotating generated table column quality.}
    \label{fig:app-annot-interface}
\end{figure*}

The annotation interface used was created using Streamlit\footnote{\url{https://streamlit.io/}} and can be found in Figure \ref{fig:app-annot-interface}.

\section{Qualitative Error Analysis}
\label{appendix:qual-error-analysis}
For each of the error types listed in Table \ref{tab:errors-schema}, we include a generated table that illustrates the error and a reference table when appropriate.
\begin{enumerate}
    \item \textbf{Different Topics} see Table \ref{tab:appendix-errors-topics}
    \item \textbf{Different Granularity} see Table \ref{tab:appenidx-errors-granularity}
    \item \textbf{Complex aspects} see Table \ref{tab:appendix-errors-complex}
    \item \textbf{Overly Specific} see Table \ref{tab:appendix-errors-specific}
\end{enumerate}

\begin{table*}[ht!]
    \small
    \raggedright
    \begin{tabular}{@{}rlrl}
    \textbf{Reference:} &&&\\
    \toprule
             & \textbf{Tasks}        & \textbf{\# categories}  & \textbf{evaluation metric} \\
        \midrule
        765adbf       & fine-grained & 100            & mean accuracy\\
         4fe680c & face & 9,131  & - \\
         \bottomrule
    \end{tabular}
    \begin{tabular}{@{}rlll}
    &&&\\
    \textbf{Generated:} &&&\\
    \toprule
    \textbf{} & \textbf{Source dataset} & \textbf{Target dataset} & \textbf{Number of images} \\ \midrule
    765adbf   & FGVC-Aircraft            & FGVC-Aircraft           & 10,000 images of airplanes \\
    4fe680c   & VGGFace2                 & VGGFace2                & 3.31 million images        \\
    \bottomrule
    \end{tabular}
    \caption{\textbf{Different Topics}: Reference table (top), Predicted table (bottom). We can see that our system generates a different (and redundant) set of aspects compared to the reference.}
    \label{tab:appendix-errors-topics}
\end{table*}

\begin{table*}[ht!]
\raggedright
\small
\begin{tabular}{@{}rlllccccccc}
\textbf{Reference}: &&&&&&&&&&\\
\toprule
\textbf{} & \textbf{Dataset} & \textbf{Year} & \textbf{Data size} & \textbf{Image} & \textbf{Text} & \textbf{Tags} & \textbf{Video} & \textbf{Audio} & \textbf{3D Model} \\
\midrule
3919117 & Twitter100K & 2018 & 100,000 & $\checkmark$ & $\checkmark$ & - & - & - & - \\
15514398 & Xmedia & 2018 & 12,000 & $\checkmark$ & $\checkmark$ & - & $\checkmark$ & $\checkmark$ & $\checkmark$ \\
\bottomrule
\end{tabular}

\begin{tabular}{@{}rlll}
&&&\\
\textbf{Generated}: &&&\\
\toprule
 &  \textbf{Dataset Name} & \textbf{Dataset Size} & \textbf{Data Types} \\
\midrule
3919117 & Twitter100k & 100,000 image-text pairs & LDA, Bag-of-Word (BoW), ... \\
15514398  & XMedia, Wikipedia, ... & 12,000 media instances & text, image, video, audio, 3D model \\
\bottomrule
\end{tabular}
\caption{\textbf{Different Granularities}: Reference table (top), Predicted table (bottom). Some aspects are removed from each table to highlight the difference in granularity. The reference table separately splits out the various data types while the generated one has a single ``Data Types'' column.}
\label{tab:appenidx-errors-granularity}
\end{table*}

\begin{table*}
    \raggedright
    \small
    \begin{tabular}{@{}rlllll}
    \textbf{Reference:} & & & & & \\
    \toprule
    \textbf{} & \textbf{Classes} & \textbf{Signer} & \textbf{Videos} & \textbf{Videos per Class} & \textbf{Controlled} \\ \midrule
    64745485  & 64               & 10              & 3200            & 50                       & $\checkmark$          \\
    54446047  & 1000             & 11-45           & 25513           & 25                       & \texttimes          \\ \bottomrule
    \end{tabular}

    \begin{tabular}{@{}rlll}
    & & & \\
    \textbf{Generated:} & & & \\
    \toprule
     & \textbf{Sign Language} & \textbf{Dataset Size} & \begin{tabular}[c]{@{}c@{}} \textbf{Number of} \\ \textbf{Subjects} \end{tabular}\\
    \midrule
    64745485 & 
    \begin{tabular}[l]{@{}l@{}} Dataset of Argentinian Sign \\ Language (LSA) presented \end{tabular}
     & \begin{tabular}[l]{@{}l@{}} 3200 videos, 64 LSA signs, \\ 10 subjects \end{tabular}  & 10 subjects  \\
    54446047 & Large-scale sign language dataset created & over 25,000 annotated videos & 222 subjects \\
    \bottomrule
    \end{tabular}
    \caption{\textbf{Complex Aspects}: Reference table (top), Predicted table (bottom).}
    \label{tab:appendix-errors-complex}
\end{table*}

\begin{table*}
    \raggedright
    \small
    
    \begin{tabular}{@{}rllll}
    \textbf{Generated:} &&&& \\
\toprule
 & \textbf{Hate Speech Dataset} & \textbf{Misinformation Dataset} & \textbf{Number of Examples}\\
\midrule
253018764 & Mentions [...] Hate Speech Dataset & N/A & multiple hate speech datasets \\
10326133 & N/A & Introduces LIAR dataset... & 12,836 \\
\bottomrule
\end{tabular}
    \caption{\textbf{Overly Specific}: The table shown is the predicted table. Note that the aspects ``Hate Speech Dataset'' and ``Misinformation Dataset'' only apply to a single paper each.}
    \label{tab:appendix-errors-specific}
\end{table*}

\end{document}